\documentclass{article}

\PassOptionsToPackage{authoryear,round}{natbib}

\usepackage[preprint]{neurips_2023}

\usepackage[utf8]{inputenc} %
\usepackage[T1]{fontenc}    %
\usepackage{hyperref}       %
\usepackage{url}            %
\usepackage{booktabs}       %
\usepackage{amsfonts}       %
\usepackage{nicefrac}       %
\usepackage{xcolor}         %
\usepackage{enumitem}
\usepackage{forloop}
\usepackage[ruled, vlined]{algorithm2e}
\usepackage{listings}
\usepackage{fvextra}

\usepackage{comment}
\usepackage{tikz}
\usetikzlibrary{shapes,arrows,decorations.pathmorphing}

\usepackage{amsmath,amsfonts,bm}

\def\eqref#1{equation~\ref{#1}}

\def\1{\bm{1}}

\DeclareMathAlphabet{\mathsfit}{\encodingdefault}{\sfdefault}{m}{sl}
\SetMathAlphabet{\mathsfit}{bold}{\encodingdefault}{\sfdefault}{bx}{n}

\newcommand{\E}{\mathbb{E}}

\newcommand{\R}{\mathbb{R}}

\DeclareMathOperator*{\argmax}{arg\,max}
\DeclareMathOperator*{\argmin}{arg\,min}

\usepackage{cleveref}
\usepackage{microtype}

\usepackage{amsmath}
\usepackage{amsthm}
\usepackage{amssymb}
\usepackage{bbm}
\usepackage{mathtools}
\usepackage{mathrsfs}
\mathtoolsset{showonlyrefs}

\usepackage{graphicx}
\usepackage{subcaption}

\usepackage{hyperref}

\usepackage{theoremref}
\usepackage{wrapfig}
\usepackage{xcolor}
\definecolor{dark-red}{rgb}{0.4,0.15,0.15}
\definecolor{dark-blue}{rgb}{0,0,0.7}
\hypersetup{
    colorlinks, linkcolor={dark-blue},
    citecolor={dark-blue}, urlcolor={dark-blue}
}

\usepackage[colorinlistoftodos]{todonotes}

\SetKwProg{Fn}{Function}{}{}
\let\oldnl\nl%
\newcommand{\nonl}{\renewcommand{\nl}{\let\nl\oldnl}}%

\usepackage{multicol}
\usepackage[utf8]{inputenc} %
\usepackage[T1]{fontenc}    %
\usepackage{hyperref}       %
\usepackage{url}            %
\usepackage{booktabs}       %
\usepackage{nicefrac}       %
\usepackage[parfill]{parskip}

\usepackage{hyperref}
\hypersetup{
    colorlinks,
    citecolor=blue,
    filecolor=black,
    linkcolor=blue,
    urlcolor=blue
}

\newcommand{\parameter}{p_{\text{tunable}}}
\newcommand{\task}{p_{\text{task}}}

\newcommand{\Parameter}{P_{\text{tunable}}}
\newcommand{\Task}{P_{\text{task}}}
\newcommand{\Feedback}{F}

\definecolor{codegreen}{rgb}{0,0.6,0}
\definecolor{codegray}{rgb}{0.5,0.5,0.5}
\definecolor{codepurple}{rgb}{0.58,0,0.82}

\definecolor{backcolour}{RGB}{245,248,250}
\definecolor{emph}{RGB}{166,88,53}
\definecolor{nightblue}{RGB}{9,49,105}
\definecolor{keywords}{RGB}{207,33,46}
\definecolor{lightpurple}{RGB}{130,81,223}

\definecolor{skyblue}{RGB}{86,168,245}
\definecolor{deepblue}{rgb}{0,0,0.5}
\definecolor{deepred}{rgb}{0.6,0,0}
\definecolor{deepgreen}{rgb}{0,0.5,0}

\lstdefinestyle{mystyle}{
    backgroundcolor=\color{backcolour},    %
    commentstyle=\color{codegreen},
    keywordstyle=\color{keywords},
    stringstyle=\color{nightblue},
    basicstyle=\fontsize{7}{8}\ttfamily,
    breakatwhitespace=true,         
    breaklines=true,                 
    captionpos=b,                    
    keepspaces=true,                 
    numberstyle=\tiny\color{codegray},
    numbersep=2pt,                  
    showspaces=false,                
    showstringspaces=false,
    showtabs=false,                  
    tabsize=2,
    emph={system,user,task\_description},
    emphstyle={\color{codepurple}},
    linewidth=1\columnwidth,
    frame=tb,    
    xrightmargin=0pt,
    xleftmargin=0.23cm,
    numbers=left,
    aboveskip=0.2cm,
    belowskip=0.1cm,
}

\title{The Importance of Directional Feedback\\ for LLM-based Optimizers}

\author{%
  Allen Nie \\
  Stanford University \\
  \texttt{anie@stanford.edu} \\
  \And
  Ching-An Cheng \\
  Microsoft Research \\
  \texttt{chinganc@microsoft.com} \\
  \AND
  Andrey Kolobov \\
  Microsoft Research \\
  \texttt{akolobov@microsoft.com} \\
  \And
  Adith Swaminathan \\
  Microsoft Research \\
  \texttt{adswamin@microsoft.com} \\
}

\begin{document}

\maketitle

\begin{abstract}

We study the potential of using large language models (LLMs) as an interactive 
optimizer for solving maximization problems in a text space using natural language and numerical feedback. 
Inspired by the classical optimization literature, we classify the natural language feedback into directional and non-directional, where the former is a generalization of the first-order feedback to the natural language space.
We find that LLMs are especially capable of optimization when they are provided with {directional feedback}.  
Based on this insight, we design a new LLM-based optimizer that synthesizes directional feedback from the historical optimization trace to achieve reliable improvement over iterations.
Empirically, we show our LLM-based optimizer is more stable and efficient in solving optimization problems, from maximizing mathematical functions to optimizing prompts for writing poems, compared with existing techniques.

\end{abstract}

\section{Introduction}

Owing to their capability to produce a diverse range of outputs similar to that of humans,
large language models (LLMs) are a powerful component for solving many difficult problems involving language, including planning~\citep{ahn2022can}, interacting with users, understanding documents~\citep{kojima2023large}, and producing executable code~\citep{gur2023real}. %
In addition to harnessing LLMs in these \emph{generative} roles, several recent works have used LLMs for \emph{optimization}. So far, these efforts, such as APO~\citep{pryzant2023automatic} and OPRO~\citep{yang2023large}, have focused on optimization of a very specific kind -- employing LLMs to produce prompts that improve (another) LLM's performance. %
In this work, we argue that LLMs' potential extends much further, to general optimization problems. We showcase that LLMs are capable of optimizing entities as dissimilar as mathematical functions and poems if they are provided with \emph{directional feedback}.

The notions of directional and non-directional feedback arise naturally in many interactive decision-making domains and are tied to the classical optimization literature~\citep{boyd2004convex}. Typically, a numerical optimizer iterates over two steps. The first step aims to %
identify a ``search direction'' for improvement.  %
This information is provided to the optimizer by an oracle, oftentimes a first-order oracle, and can be viewed as directional feedback.
The second step decides what to change about the input. The applicability of various optimization methods depends on whether the directional feedback information is available or not. Scenarios without directional feedback are confined to black-box optimization methods such as evolutionary search~\citep{mitchell1998introduction}, Bayesian optimization~\citep{mockus1998application}, or policy gradient~\citep{sutton1999policy}. However, when the directional feedback is available, one can choose the much more efficient gradient-based optimization method, such as stochastic gradient descent or exact line search~\citep{boyd2004convex}. This insight motivates our use of directional feedback in LLM-driven optimization.

As we show, the presence or absence of directional feedback and the possibility to access them is crucial for LLM-based optimization. For a systematic study of factors that make an optimization process challenging, we choose one of the difficult tasks proposed in the work on OPRO (listed as failure cases in Appendix A of OPRO~\citep{yang2023large}) -- navigating a bumpy loss function landscape. We discover that an LLM-based optimizer's performance varies with the type of information the feedback carries, and, given proper feedback, LLMs can strategically improve over past outputs, which makes this previously unsolvable task solvable. In addition, we demonstrate that 
using LLMs to ``synthesize'' feedback from a history of observations and prompts can help optimization too.

We also explore LLMs' optimization potential in a completely different setting. Given the importance of feedback type in the LLM-based optimization process and the lack of benchmarks that generate verbal feedback automatically, we
create a synthetic poem writing environment, where one can programmatically create feedback for the LLMs. The poem environment is a family of tasks where an LLM is asked to write a poem. A distinguishing feature of this benchmark is that the poems must satisfy some constraints, such as the number of syllables per line. By leveraging and synthesizing feedback, we show that an LLM can sequentially optimize a poem-generation prompt to yield a high success rate of producing constraint-satisfying poems. 
Our results highlight the importance of studying the role of feedback in the broader LLM-based text optimization landscape.

\section{Preliminaries: Prompt Optimization for LLM-based Agent}

An LLM-based agent's behavior is modulated through the prompts used as inputs to the LLM. We describe the interactive decision-making problem encountered by an LLM-based agent, and how prompt optimization through interactive feedback can improve the agent over time. In the following, uppercase letters, e.g. $X$, denote random variables or sets. Lowercase letters denote realizations of the random variables or set elements, e.g. ``$X = x$" states that a r.v. $X$ takes on value $x$. Greek letters, e.g., $\xi$, denote parameters indexing probability distributions.

Consider an agent encountering a complex task such as generating a poem with logical constraints. The task is communicated to the agent via a text prompt $p_{\text{task}} = \textit{``Generate a 5-line poem with a 5-7-5-7-5 syllable pattern''}$. 
The LLM produces output text $o_1 \sim \Pr_{\tau}(O \mid p_{\text{task}}, p_{\text{tunable}})$ (e.g., a poem, or plans, or executable code, or other texts as prompted), where $\tau$ captures LLM hyper-parameters like sampling temperature. $p_{\text{tunable}}$ contains any orchestrated text inputs from other modules surrounding the LLM. Shortly, we will develop modules that will incorporate information gathered over time to update $p_{\text{tunable}}$. By analogy of tunable parameters from an ML model, we view $p_{\text{tunable}}$ as the tunable ``parameters'' of an LLM-based agent.

Based on the generated output $o_1$, a scalar reward $r_1$ and optionally feedback $f_1$ are generated from the environment (e.g., human user response, or logs generated by executing code in a programming environment) and passed to the agent. For ease of notation, we assume $r \sim R$ and $f \sim F$, but we do not make specific assumptions of the underlying distributions. The reward can be a task success or failure boolean from the environment, or user-provided thumbs-up/down signal. This interaction process iterates $o_1 \rightsquigarrow \{r_1,f_1\}, \dots , o_t \rightsquigarrow \{r_t,f_t\}$, until the environment terminates the interaction session. 
Figure~\ref{fig:llm_schematic} illustrates the interaction process; for example, in Minecraft Voyager~\citep{wang2023voyager}, a prompt $p_{\text{task}}=\text{``Build a house''}$ is translated into a code-generation request using internal orchestration that prepends a specific $p_{\text{tunable}}$\footnote{In Voyager, these prompts are hand-engineered and not automatically tuned.}. The produced code $o_t$ is executed in the Minecraft environment to generate error/debug/return messages $f_t$ as well as task completion flag $r_t$ that are returned to Voyager to refine the code $o_{t+1}$ in subsequent iterations. 
The interaction session ends when the user prompting the Voyager agent terminates it. Note that $p_{\text{task}}$ can be interactively updated within a session (e.g., user providing additional hints or rephrasing the task), and we only assume that the rewards and feedbacks observed are consistent with the task that the agent is prompted to solve.

\begin{figure}[htb]
    \centering
    \begin{subfigure}[b]{0.5\textwidth}
    \centering
    \scalebox{0.6}{%
    \begin{tikzpicture}[auto,node distance=2.5cm,>=latex',font=\Large]

    \node [draw, rectangle, text width=2cm, minimum height=1.5cm] (llm) {LLM Agent};
    
    \node [left of=llm, text width=1.5cm, align=center] (taskprompt) {Task Prompt $p_{\text{task}}$};
    \node [above of=llm, text width=1.5cm, align=center] (tunableprompt) {Tunable Prompt $p_{\text{tunable}}$};
    
    \node [right of=llm, text width=1.5cm, align=center] (output) {Output $o$};
    
    \draw[->] (taskprompt) -- (llm);
    \draw[->] (tunableprompt) -- (llm);
    \draw[->] (llm) -- (output);
    
    \draw[decorate, decoration={snake, amplitude=0.5mm, segment length=2mm, post length=1mm}] (output) -- ++(1.5cm,0); 
    \node [right of=output, text width=1.5cm, align=center] (feedback) {Feedback $f$};
    \node [above=0.0cm of feedback, text width=1.5cm, align=center] (reward) {Reward $r$};
    
    \end{tikzpicture}
    }
    \caption{Schematic of LLM-based agent. $p_{tunable}$ can be updated from feedback and/or previous experiences via our sequential optimization.}
    \label{fig:llm_schematic}
    \end{subfigure}
    \hfill
    \begin{subfigure}[b]{0.4\textwidth}
    \centering
    \scalebox{0.6}{%
    \begin{tikzpicture}[auto, node distance=2.5cm,>=latex',font=\Large]

    \node [draw, rectangle, text width=2cm, minimum height=1.5cm] (llm) {LLM-Optimizer};
    
    \node [left of=llm, text width=1.5cm, align=center] (taskprompt) {Optimizer Prompt $p_{\text{optimizer}}$};
    \node [above of=llm, text width=2cm, align=center] (tunableprompt) {Interaction History $\{ o_t, r_t, f_t \}$};
    
    \node [right of=llm, text width=1.5cm, align=center] (output) {Tuned prompt ${p'}_{\text{tunable}}$};
    
    \draw[->] (taskprompt) -- (llm);
    \draw[->] (tunableprompt) -- (llm);
    \draw[->] (llm) -- (output);
    \end{tikzpicture}
    }
    \caption{LLM-based Optimizer is a specific LLM-based agent that incorporates previous experiences into the tunable prompt $p_{\text{tunable}}$ of another LLM-based agent.}
    \label{fig:llm_optimizer}
    \end{subfigure}
\end{figure}

Define an LLM agent as $\pi: \Task \times \Parameter \rightarrow O$. The distribution $O$ is defined by $\parameter$ alone, which we regard as the parameter of the LLM agent. The optimization problem we need to solve is to find $\parameter^\star \coloneqq \argmax_{\parameter} \E_{o} \left[ r \mid \pi(\task, \parameter) \right]$. We can define an optimizer $g: \Parameter \times \Feedback \times R \rightarrow \Parameter$, where the goal is to find $\parameter^\star$ through a limited number of times that $\pi$ attempts the task. An optimal optimizer $g^\star$ can find $\parameter^\star$ with the fewest number of attempts.

\section{Optimizing LLM-based Agents}

The LLM-based Optimizer is a specific instance of an LLM-based agent. It can be used to improve another LLM-based agent using collected experience so that the generated outputs have higher expected reward $\mathbb{E}_{o} \left[ r \mid o, p_{\text{task}} \right]$. The LLM-based optimizer takes a collection of Output-Reward-Feedback $(o,r,f)$ tuples via its tunable prompt (see Figure~\ref{fig:llm_optimizer}), and is tasked with generating a prompt ${p'}_{\text{tunable}}$ for the LLM-based agent.

\subsection{Fundamentals of LLM Optimization \label{sec:fund}}

The most common approach for optimization is through an iterative solver that improves monotonically. However, in order to construct an iterative solver, the optimization problem needs to satisfy a few assumptions.
To establish intuitions, we start with numerical optimization in a function approximation-based supervised learning setting. Given a hypothesis $h$ and data sample $(x, y)$, let $\tilde y = h_\theta(x)$. With a loss function $\ell: X \times Y \times \Theta \rightarrow \R$, we can define $L(\theta) = \E_{(x, y)} \left[\ell(\theta, x, y) \right]$. The goal is to find $\theta^\star = \argmin_\theta L(\theta)$. 
To achieve this goal, a valid optimization procedure proposes a new $\theta$ for $k$ number of times. They usually consist of two steps:
\begin{enumerate}[label=\textbf{S\arabic*},leftmargin=*,start=1]
    \item \textbf{Finding Valid Search Direction}: We need to find useful information, such as a descent direction $\Delta \theta^{(k)}$ that can inform the update step. The usefulness of the information is tightly coupled with what the update step is.
    \item \textbf{Decide Update Rules}: We need to decide how to update $\theta$. A typical update procedure is simply: $\theta^{(k+1)} = \theta^{(k)} + t^{(k)} \Delta \theta^{(k)}$, if $\Delta \theta^{(k)}$ is informative, where $t_k$ is the step size.
\end{enumerate}
If $L$ is convex, then the criteria to determine whether $\Delta \theta$ is informative is quite simple: we can use the gradient of $L$. From convexity, we know that $\nabla  L(\theta^{(k)})^T (\theta^{(k+1)} - \theta^{(k)}) \geq 0$ implies $L(\theta^{(k+1)}) \geq L(\theta^{(k)})$. Then we can set the descend direction $\Delta \theta^{(k)}$ so as to satisfy $-\nabla  L(\theta^{(k)})^T \Delta \theta^{(k)} \ge 0$. A simple way to satisfy this criterion is let $\Delta \theta^{(k)} \coloneqq - \nabla L(\theta^{(k)})$, which is the gradient descent method (GD). However, more complicated update rules can be used, such as backtracking line search~\citep{boyd2004convex}. Note that we do not always need to satisfy \textbf{S2}. For example, in an evolutionary search algorithm, many candidates are proposed and the update rule simply keeps the candidate with the best score.

Then we can contrast the setting with an LLM optimization problem. If we want to have an iterative descent algorithm to find the optimal prompt for an LLM agent, then we need to consider the following properties:
\begin{enumerate}[label=\textbf{S\arabic*},leftmargin=*]
    \item \textbf{Search Direction}: We should obtain useful information, analogous to $\nabla L(\theta)$, to help inform the optimizer on how to update the parameter $\parameter$.
    \item \textbf{Update Parameter}: Unlike the numerical case, where basic algebra can be applied to update parameters, it is unclear whether there is a predefined notion of $\Delta \parameter^{(k)}$ in text space. This distance $\Delta \parameter^{(k)}$ in the best case, can be assessed by human intuition over the semantics of the text, in the worst case, can be completely arbitrary.
\end{enumerate}

To propose an algorithm using LLM as an optimizer, we make the following assumptions:
\begin{enumerate}[label=\textbf{A\arabic*},leftmargin=*]
    \item \textbf{Permissible Search Direction}: There exists useful information, which we describe as feedback, $f$, for an LLM-based optimizer $g$ such that $g$ can propose a $\parameter^{k+1}$ where $\E_{o} \left[ r \mid \pi(\task, \parameter^{k+1}) \right] \geq \E_{o} \left[ r \mid \pi(\task, \parameter^{k}) \right]$.
    \item \textbf{Valid Update}: The LLM-based optimizer $g$ can modify $\parameter$ based on $f$, where direction of change: $\Delta \parameter$ is determined by the information contained in $f$ (i.e., not a random text edit).
\end{enumerate}

In the next few sections, we describe a few possible settings where these assumptions can be satisfied or need not be. We assume $\textbf{A2}$ is always satisfied. The first setting we discuss is that it is possible that LLM itself acts as a black-box optimizer. For example, it might obtain a valid search direction by implicitly computing finite differences between inputs and outputs in text space.

\paragraph{LLM Might Implicitly Perform Newton's Method}
 Similar to Newton's Method using finite differences as an approximate gradient, perhaps the LLM can implicitly compute the following function: 
 \begin{align}
     \nabla R = \lim_{\Delta \parameter^{(k)} \rightarrow 0} \frac{\E_{o} \left[ r \mid \pi(\task, \parameter^{k}) \right] - \E_{o} \left[ r \mid \pi(\task, \parameter^{k+1}) \right]}{\Delta \parameter^{(k)}}
 \end{align}
If we think this is possible, then the input to the LLM can be tuples of $(\parameter^1, r_1), ..., (\parameter^k, r_k)$. Although it is unclear if this is truly the case, this shows that the optimizer needs to retain a history of how past prompts $p$ have influenced the reward $r$.

\paragraph{Feedback Can Be Directional} The other possibility, not relying on the black box ``magic'' of the LLM's internal process, is to hope that somehow a permissible search direction $f$ is given to us from an external source. Humans give directional feedback quite often: ``This coffee is too hot for me.'' or ``Can you lower the room temperature?'' The first feedback implicitly asks the agent to make a cooler coffee (but not saying exactly how cool it should be). The second feedback asks the agent to turn down the room temperature (but without specifying which temperature to set). Imagine if the agent's action (output space $O$) for both cases is to write API calls to set temperature; then we know immediately what $\Delta O$ should be -- keep everything the same, but enter a lower temperature value. After the adjustment, a user might say: ``This coffee is now too cold for me.'' or ``I'm freezing!'' Making this kind of feedback very similar to the gradient information we get from numerical optimization. This suggests that, in some cases, incorporating feedback (or somehow obtaining directional feedback) could be helpful for the optimization procedure. In such cases, we can provide the LLM optimizer with examples of $(\parameter^1, f_1, r_1), (\parameter^2, f_2, r_2), ..., (\parameter^k, f_k, r_k)$ (i.e. additionally providing the feedback in each round to the optimizer).

\paragraph{Non-directional Feedback} There is another type of feedback we can consider. This type of feedback contains useful information but is not directional because they do not directly inform us how to change the input $O$. For example, feedback like ``I can't drink this coffee because the temperature is not quite right.'' This feedback clearly states that the attribute of ``temperature'' is important to the user and is not satisfactory. However, it does not tell us whether we should make a coffee that's hotter or colder. Coffee can have many attributes, such as ``temperature'', ``acidity'', ``roast'', ``sweetness'', or ``cream-level''. This feedback is more useful than a scalar reward because it \emph{explains} attributes that affect $R$ and allows us to focus more on a single dimension of many attributes.

\paragraph{Reward as Feedback / No Feedback} Unlike directional and non-directional feedback, which usually contains information about how to change $\parameter$, score-based feedback only gives back a numerical value indicating how well $\parameter$ performs. In our setup, this means LLM-Optimizer only observes reward $r$ without $f$.

\subsection{Sequential Prompt Optimization}
\label{sec:spo_alg}

Inspired by the descent method in numerical optimization, we propose an algorithm that aims to satisfy the requirement of descent methods such that we can reach the extremum. We define the following optimization process with an LLM-based agent $\pi$. With an initial tunable prompt $\parameter$ and a task description $\task$ from the environment, agent $\pi$ samples an output $o_1$. The environment returns a reward $r_1$ and feedback $f_1$. An LLM-optimizer stores $(o_1, r_1, f_1, \parameter)$ in a history buffer $\mathbb{H}$. When the buffer becomes large, we subsample $H$ from $\mathbb{H}$. The LLM-optimizer proposes a new tunable parameter $\parameter'$. We make an explicit decision on whether to replace $\parameter$ with $\parameter'$ based on the reward evaluated on the distribution $o$ and $o'$. The full procedure is in Algorithm~\ref{alg:spo}. We describe the implementation choices for each component of our iterative solver below.

\paragraph{Agent} Agent is an LLM that takes in a tunable instruction prompt $\parameter$, description of task $\task$ and produces an output $o$. Unlike previous work, it does not have access to the history of interactions it had with the environment. This design decision gives the optimizer a higher degree of control over the agent's behavior. Whether the agent should have access to its own history or what history to access should be determined by the optimizer, not by itself. It is different from some other existing work. For example, Voyager~\citep{wang2023voyager} would allow the agent to see all of its interaction histories (and errors they make). React~\citep{yao2022react} also allows the agent to see the full error trace. Allowing the agent to see its past errors is a specific design choice on $\parameter$. We let the optimizer decide what $\parameter$ should be (and its decision space includes the human-engineered choices in prior works). %

\paragraph{Prompt Proposal} We define the prompt proposal module as $\Delta: \Task \times \Parameter \times F \times R \rightarrow \Parameter$. This module looks at the task description, past prompts, and the feedback each prompt receives and proposes a new prompt. If the environment provides directional or non-directional feedback, this module should take in $F$ as well. Even though past prompts were included, this module is allowed to produce completely new prompts. 

\setcounter{algocf}{1}
\noindent\makebox[\textwidth][c]{
\begin{minipage}{0.6\textwidth}
\centering
\begin{algorithm*}[H]
\SetAlgoLined
\SetKwFunction{sample}{Sample}
\KwIn{Given $s, R$, an LLM-based agent $\pi$, an LLM-based prompt proposal module $\Delta$, $\parameter^0$, $\task$, and $K$ iterations.}
\KwOut{$\parameter^{K}$}
\BlankLine
$\mathbb{H} = \emptyset$ \\
\For{$k \leftarrow 0...K$}{
    $o_k, r_k \sim \pi(\task, \parameter^{(k)}), R$ \\
    $f_k \sim F$ or $\hat F(\task, o_k, r_k)$ \\
    $H = \sample(\mathbb{H})$ \\
    $p^{(k+1)} = \Delta(H, \{\task, \parameter^{k}, f_k, r_k\})$ \\
   \If{$\E_{o} \left[ r \mid \pi(\task, \parameter^{k+1}) \right] \geq \E_{o} \left[ r \mid \pi(\task, \parameter^{k}) \right]$}{ 
     $\parameter^{k+1} = \parameter^{k}$
    }
    $\mathbb{H} = \mathbb{H} \cup \{o_k, r_k, f_k, \parameter^{k}\}$ \\
}
 \Return{$\parameter^{K}$}
 \caption{Sequential Prompt Optimization}
 \label{alg:spo}
\end{algorithm*}
\end{minipage}
}

\paragraph{Feedback Synthesizer} If numerical feedback is the only type of feedback given, we can design a feedback synthesizing module $\hat F: \Task \times O \times R \rightarrow F$. It takes in $(o_1, r_1), ..., (o_k, r_k)$ and produces feedback to the prompt proposal module. We designed this module to ask the question, ``How should the input be changed to have a greater effect on the objective/output?'' We note that we more specifically prompt the LLM to think about the difference in the input space that would impact the difference in output space, different from APO, where they ask the LLM to ``give reasons why the prompt could have gotten these examples wrong.''

\paragraph{Prompt Selector} In order to satisfy the descent method assumption \textbf{A1} (permissible search direction), we need to guarantee that $\parameter'$ is an improvement over $\parameter$. This can be achieved by setting a selection criterion that requires $p^{(k+1)}$ to get a higher reward than $p^{(k)}$. A simple criterion is to improve the average performance over the distribution of $O$: $\E_{o} \left[ r \mid \pi(\task, \parameter^{k+1}) \right] \geq \E_{o} \left[ r \mid \pi(\task, \parameter^{k}) \right]$.

\section{Experiments}

\begin{figure}[t]
    \centering
    \includegraphics[scale=0.7]{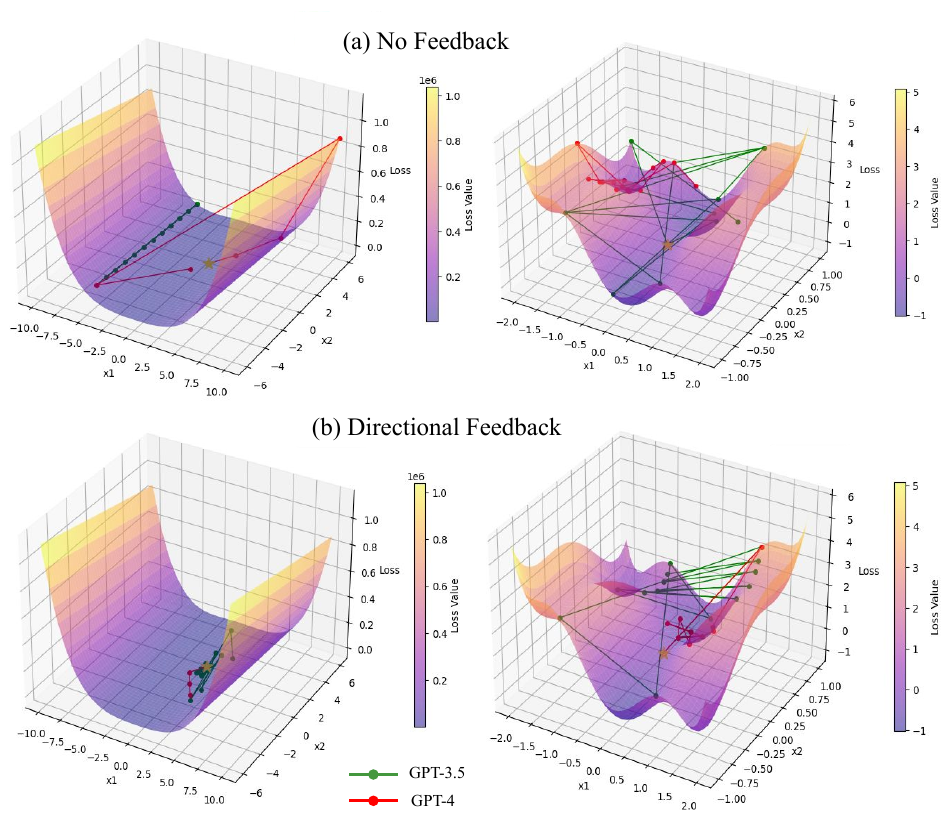}
    \caption{We visualize the optimization trajectory path made by the Optimizer Agent with GPT-3.5 and GPT-4. The loss landscape on the left is the Rosenbrock Function, and on the right is the Six-Hump Camel Function. }
    \label{fig:loss_traj}
\end{figure}

\subsection{Numerical Optimization}
\label{sec:expt_loss_opt}
We test if LLM can possibly do optimization and what are the necessary ingredients for it to find the optimal solution in an optimization problem. We set up this task as a more controlled study before studying prompt optimization. In prompt optimization, it is often hard to know what the optimal prompt or a good search direction is. With a numerical optimization problem instead, both the optimal solution and a good search direction are well-defined.
We use a set of classic optimization problems\footnote{\url{https://www.sfu.ca/~ssurjano/optimization.html}} that require LLMs to find $x$, a 2-dimensional vector. The implementation used in this paper is included in LLF-Bench~\citep{cheng2023llf}.
\begin{enumerate}[leftmargin=*]
    \item Task: Given a task description $\task$ and a function $J$ (which is hidden from the LLM), we sample a random starting point $(x_0, J(x_0))$, $x_0 \sim X$. An LLM is asked to produce $x$ to minimize $J$.
    \item Optimizable variable: $X$. The LLM is asked to output $x$ directly. Here, $\parameter$ is the same as $x$.
    \item Output process $O$: the output module takes $x$ and directly outputs $x$, an identity function. %
    \item Reward $R$: $R(x) = -J(x)$.
    \item Feedback $F$:
    \begin{itemize}
        \item Directional Feedback: $\nabla R(x) = \frac{dR}{dx}$, the first-order derivative of the output.
        \item Non-directional Feedback: We compare the partial derivatives $\frac{\partial R}{\partial x_1}$ and $\frac{\partial R}{\partial x_2}$, and tell the LLM which dimension of $x$ should be changed to accomplish the task (but without telling LLM in which direction to change).
    \end{itemize}
\end{enumerate}

This experiment does not use the full setup of Algorithm~\ref{alg:spo}. We only test our LLM-based optimizer $\Delta$ and our feedback synthesizer $\hat F$. We choose four functions: Booth, McCormick, Rosenbrock, and Six-Hump-Camel Function. They were chosen because the optimal $x$ that minimizes these functions is not $[0, 0]$. In our initial experiments, we found that the LLM is quick to guess $[0, 0]$ for any problem, which trivializes the optimization problems where the function attains its minimum at $[0, 0]$.

We define simple regret as $\text{Reg}(\Delta) = |J(x_T) - J(x^\star)|$, where $J$ is the function we try to minimize and $T$ is the number of optimization steps we allow the optimizer $\Delta$ to take. We define cumulative regret as $\text{CuReg}(\Delta) = \sum_{t=1}^T  |J(x_t) - J(x^\star)|$. Intuitively, simple regret corresponds to how close is an optimizer's final answer $x_T$ to the correct answer $x^\star$. Cumulative regret describes how ``efficient'' is the optimizer at finding the $x^\star$. We compare three models: $\Delta$ with \textbf{GPT-3.5}, \textbf{GPT-4}, and a stochastic gradient descent algorithm (\textbf{SGD}) with a small yet fixed learning rate. 
We list all of the prompts we used and additional experimental details in Appendix~\ref{app:loss_opt}. 
In the reported results, we run 10 trials and allow $\Delta$ to take at most 10 optimization steps.

\begin{figure}[t]
    \centering
    \includegraphics[width=\textwidth]{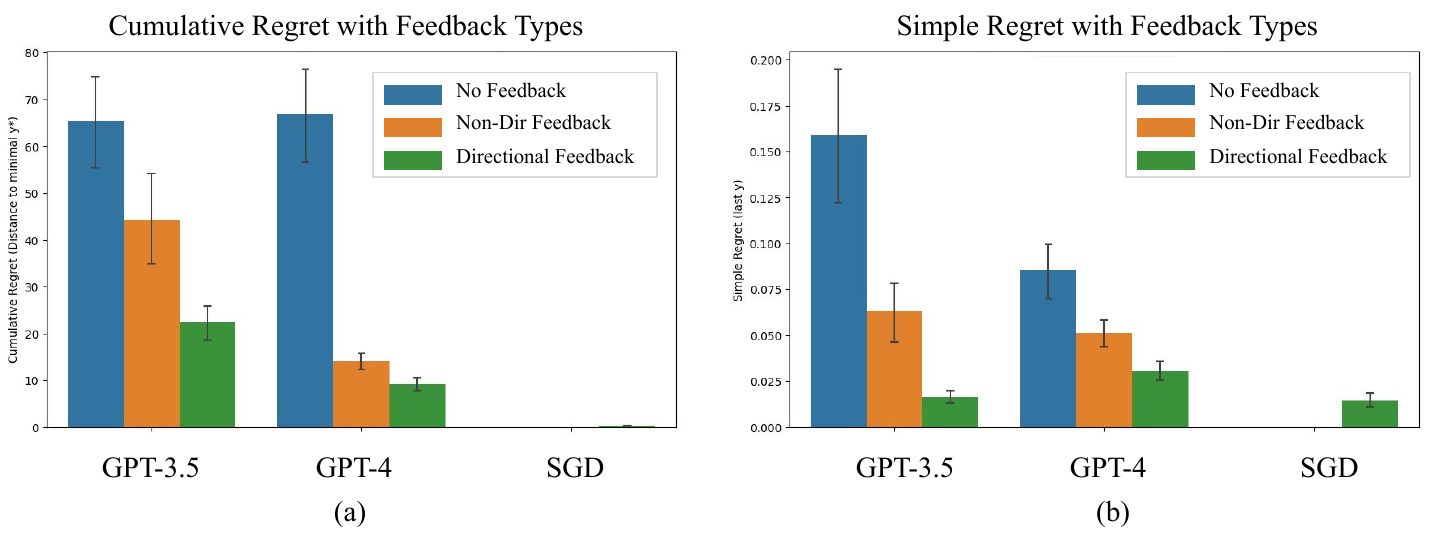}
    \caption{We plot the average Cumulative Regret and Simple Regret of each condition over 10 trials. Each algorithm is allowed to take 10 steps. We tuned the SGD learning rate to ensure that it was not too large or too small. The result is aggregated over 4 loss functions.}
    \label{fig:regret_hist}
\end{figure}

\begin{enumerate}[label=\textbf{RQ}\arabic*,leftmargin=*]
    \item Can an LLM Implicitly Perform Newton’s Method, given $(x_1, J(x_1), ..., (x_k, J(x_k))$?
\end{enumerate}

From Figure~\ref{fig:loss_traj}, we can see that LLM, as an optimizer, has a rough sense of direction given a history of past explorations. In Figure~\ref{fig:loss_traj} (a), we note that in both loss landscapes, although GPT-3.5 often fails to find the minimal point without feedback (green lines), GPT-4 is able to understand the past history and make new proposals of $x$ that incrementally minimizes $J(x)$. This suggests that even though there is no explicit gradient computation, LLM can be asked to ``improve'' based on a history of observations.

\begin{enumerate}[label=\textbf{RQ}\arabic*,start=2,leftmargin=*]
    \item Does directional Feedback help the optimization process? Do other types of feedback help as much?
\end{enumerate}

We designed the prompt space for the LLM-based optimizer $\Delta$ to insert feedback text right after the observation text and with an additional wording that reads, ``You should incorporate the suggestion.'' Besides this change, the optimizer agent prompt stays the same between no feedback and with feedback conditions. The full prompt is available in Appendix~\ref{app:loss_opt}. 

From Figure~\ref{fig:loss_traj} (b) and Figure~\ref{fig:regret_hist}, we can see that both GPT-3.5 and GPT-4 are able to take advantage of the additional feedback information and improve their search direction. Feedback can help both a weaker model (GPT-3.5) and a strong model (GPT-4). A stronger model can improve more, even if the feedback has less information (see the comparison between Non-directional Feedback and Directional Feedback in \Cref{sec:fund}). 

\begin{figure}[htb]
    \centering
    \includegraphics[width=\textwidth]{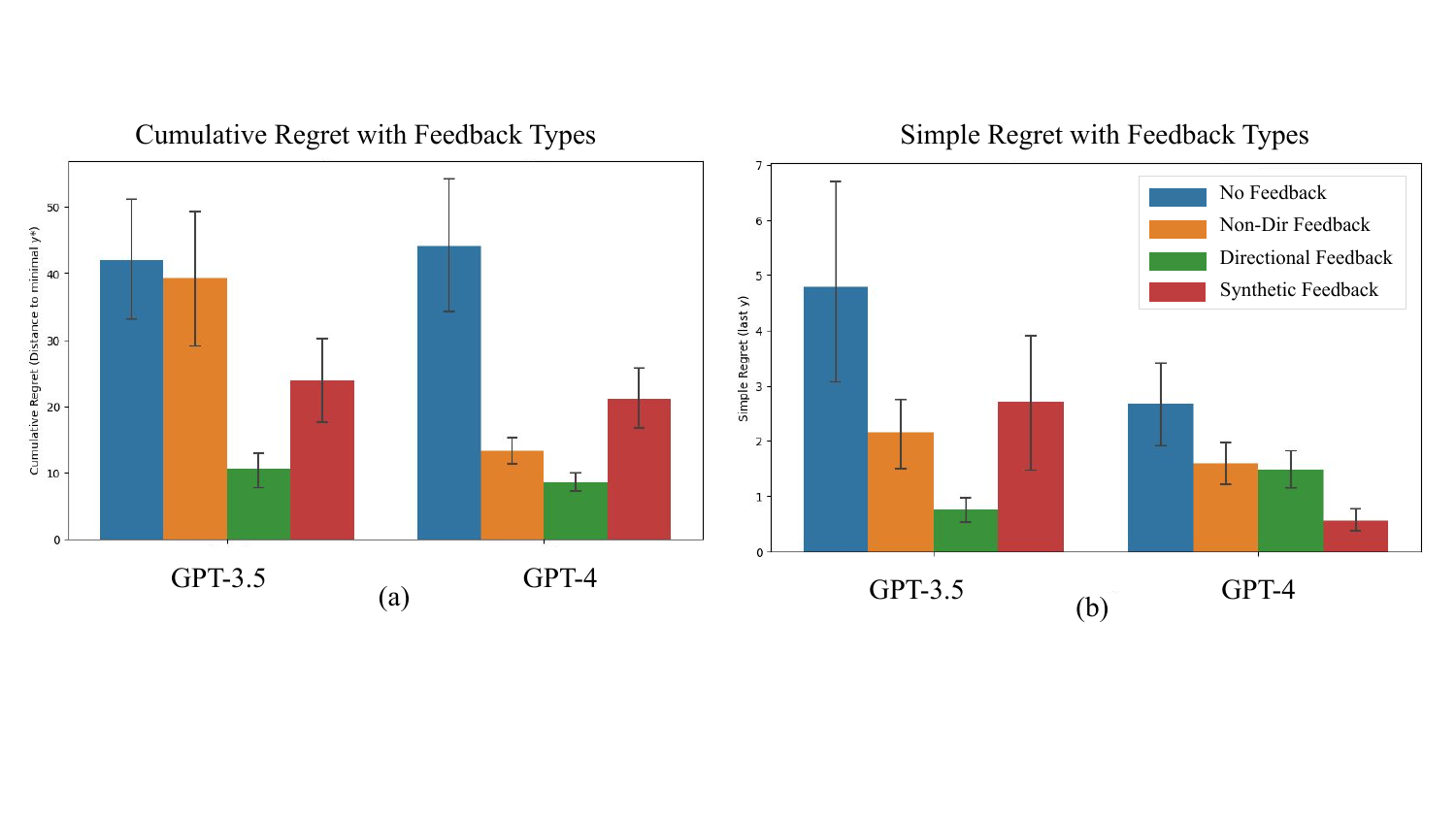}
    \vspace{-0.9in}
    \caption{We plot the average Cumulative Regret and Simple Regret of each condition over 10 trials and compare different feedback types. \textbf{Synthetic Feedback} is generated by the same LLM as the optimizer.}
    \label{fig:regret_amp_hist}
\end{figure}

Although loss minimization is a challenging task for LLMs, with some amount of feedback, LLMs are able to find a final $x$ that is similar to a classic optimization algorithm like SGD (see Figure~\ref{fig:regret_hist}b) -- the simple regret is similar across the two methods. It is worth noting that GPT-4's final proposed $x$ is not as close to the optimal as GPT-3.5. This is potentially because both models decide their own step size, and we are limiting the optimization horizon to 10 steps.

\begin{enumerate}[label=\textbf{RQ}\arabic*,start=3,leftmargin=*]
    \item If directional feedback is missing, can we replace it with an LLM module to enhance whatever feedback is available?
\end{enumerate}

Oftentimes, direct and useful feedback might be missing from the environment. In this experiment, we design a feedback synthesizer module (described in Sec~\ref{sec:spo_alg}) that takes the output from the model and the reward and synthesizes feedback to improve the next output. Different from methods such as self-reflection, self-criticism, or thinking step-by-step, the feedback synthesizer asks questions similar to ``What should I change about $x$ that will result in a larger change in $y$?'', whereas self-reflection usually asks the model to reflect upon past ``mistakes'' and critique what they did wrong.

In Figure~\ref{fig:regret_amp_hist}, we show that we can synthesize feedback from a history of past outputs and rewards that is able to guide the optimizer LLM to find a better solution. Synthesized feedback is not as informative as directional feedback that comes from the environment, but it can easily outperform settings where no feedback is given.

\begin{figure}[t]
    \centering
    \includegraphics[width=\textwidth]{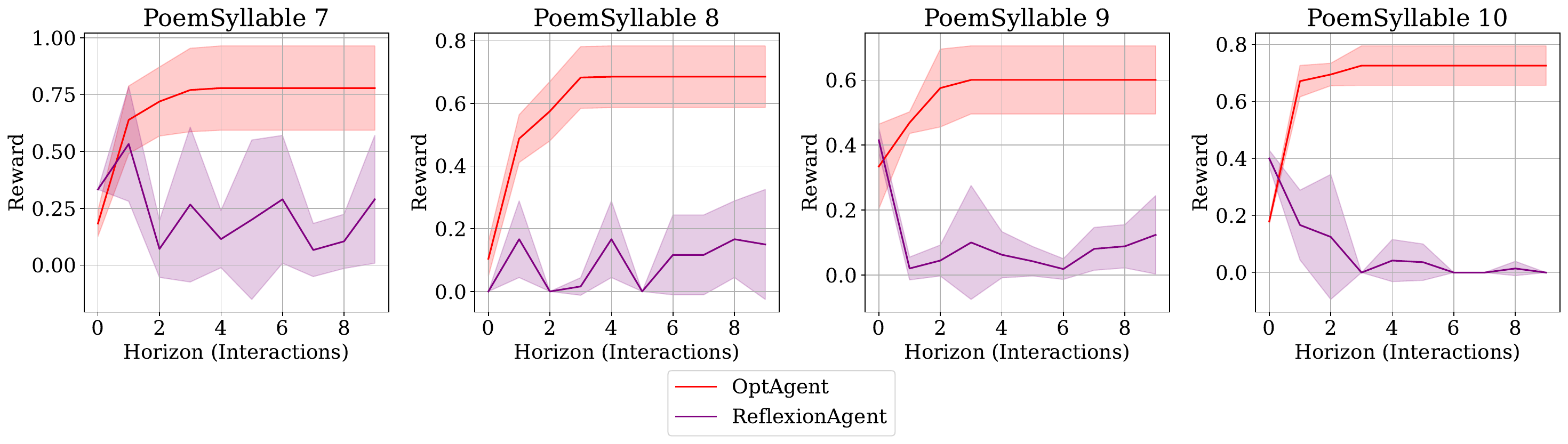}
    \caption{We show the reward for each policy after each round of interaction with the environment. OptAgent (our algorithm) is in red.}
    \label{fig:poem_single_op}
\end{figure}

\subsection{Poem Generation}

In Section~\ref{sec:expt_loss_opt}, we have validated the importance of directional feedback. We now validate our optimization setup on a more practical domain, where we have to optimize over a prompt that controls how another LLM-based agent produces output. 
Unlike a typical mathematical reasoning or language task, the optimal output $o^\star$ in the poem generation environment is defined over a text distribution. Even for a single task $\task$, we do not have direct access to the optimal distribution $o^\star$ that can obtain the highest expected reward. This is because, for a text constraint satisfaction problem, many generated texts can obtain the full reward. We discuss the implementation details below. This task is now included in LLF-Bench~\citep{cheng2023llf}.

A formal poem is a writing assignment that requires the creation of a poem to satisfy some requirements regarding its form. For example, Haiku is a type of formal poem that asks for three lines that form a 5-7-5 syllable pattern. This is a challenging task for both GPT-3.5 and GPT-4, but easy for us to verify whether the generated poem has satisfied the constraint. We can almost treat this as a code optimization problem, where we can check if each line of the poem, or the entire poem satisfies the constraint, and provide line-by-line feedback if needed. The mechanical aspect of this task makes it a perfect toy environment for prompt optimization.

\begin{enumerate}[leftmargin=*]
    \item Task: Generate a poem with a given constraint sampled from a set of constraints.
    \item Optimizable variable: $\Parameter$: This is the prompt $\parameter$ for the LLM-based agent that we want to update and optimize. 
    \item Output process $O$: the LLM agent takes the prompt $\parameter$ and follows its suggestion and a task description $\task$ to produce a poem $o$.
    \item Reward $R$: The fraction of lines in the generated poem that satisfy the constraint described by $\task$. $r \in [0, 1]$.
    \item Feedback $F$:
    \begin{itemize}
        \item Directional Feedback: We print out the number of syllables in the current line and whether LLM needs to increase or decrease the number of syllables in that line.
        \item Non-directional Feedback: We print out how many lines violate the poem writing constraints.
    \end{itemize}
\end{enumerate}

In the following experiment, we use the full setup of Algorithm~\ref{alg:spo}. We allow each agent to take 10 optimization steps. We name our agent \textbf{OptAgent}. It produces an instruction that will be sent to the poem generation agent to produce a poem. The poem-generation agent will not see the history of mistakes or any other information. We additionally evaluate \textbf{Reflexion agent}~\citep{shinn2023reflexion}. We set up four tasks: generating poems that contain 7, 8, 9, or 10 syllables for each line.

We show that in Figure~\ref{fig:poem_single_op}, we can reliably select prompts that improve the policy performance for each task. The prompt selection step in our optimization algorithm ensures monotonic improvement in rewards. %
Otherwise, it will reject the updated prompt and keep the previous prompt. This differs from the Reflexion Agent which is not guaranteed to improve after each interaction.

\section{Conclusion}

This paper argues that LLMs can successfully optimize a wide range of entities ranging from mathematical functions to prompts for textual tasks if provided with directional feedback. We empirically show on challenging numerical optimization scenarios and constrained text generation tasks that utilizing either environment-provided or synthesized feedback is a crucial piece in LLM-based optimization. We emphasize that this is an early work on general LLM-based optimizers. LLMs' potential in this role is still to be realized with new methods for directional feedback generation.

\begin{ack}
We would like to thank Ahmed Awadallah, Jennifer Neville, and Ricky Loynd for their feedback and discussions.
The work was performed between June and September of 2023 during the first author's internship at Microsoft Research in Redmond.
\end{ack}

\bibliographystyle{plainnat}
\bibliography{example_paper}

\appendix 

\section{Appendix}

\subsection{Loss Optimizing Experiment Details}
\label{app:loss_opt}

We designed the prompt for two agents. The prompt is written in a Handlebar syntax, where ``\{\{\}\}'' indicate variables to be replaced. A brief guide on this syntax is available here\footnote{\url{https://github.com/guidance-ai/guidance}}.

For the \textbf{LLM-Optimizer}:

\begin{small}
\begin{lstlisting}[language=HTML,breaklines=true,showstringspaces=false,basicstyle=\fontsize{7}{7}\ttfamily, numbers=none]
{{#system~}}
You are trying to minimize the output (y) of a function by choosing input (x).
{{~/system}}

{{#user~}}
{{task_description}}

This is what you have previously chosen for x and what the ys were:
{{observation}}

{{feedback}}

You should incorporate the suggestion to output the next x.
Please output the next x that will make this function output the smallest y.
You cannot repeat the same x, doing so will result in a penalty.

Format: x = [x1, x2]
Output:
{{~/user}}
\end{lstlisting}
\end{small}

For the \textbf{feedback synthesizing LLM}:
\begin{small}
\begin{lstlisting}[language=HTML,breaklines=true,showstringspaces=false,basicstyle=\fontsize{7}{7}\ttfamily, numbers=none]
{{#system~}}
You are trying to minimize the output (y) of a function by choosing input (x).
{{~/system}}

{{#user~}}
You are trying to minimize the output (y) of a function by choosing input (x).
You get to observe y once you choose the value of x, where x is a 2-dimensional vector.
This means x = [x1, x2], where x1 and x2 are real numbers.
The goal is to choose x such that y is as small as possible.

Here is a list of x and how it affects y:
{{#each history}}
{{this.action}}
{{this.observation}}
===================
{{~/each}}

For x = [x1, x2]
What are the suggestions you can give to the user to make y smaller?
For example, here are some of the things you can suggest:
- Changing x1 seems to have a bigger effect on y than changing x2.
- Make a larger change on x2
- Increase x1 by 1.2
- Decrease x2 by 0.5
- Try to increase x1 and decrease x2 at the same time
Or any other kind of suggestion. Do not make a suggestion that's the form of a question.
You should only make a one-sentence suggestion that's brief and short.

Suggestion:
{{~/user}}
\end{lstlisting}
\end{small}

\subsection{Poem Experiment Details}

For the \textbf{LLM-agent} that generates the poem, we use the following prompt:
\begin{small}
\begin{lstlisting}[language=HTML,breaklines=true,showstringspaces=false,basicstyle=\fontsize{7}{7}\ttfamily, numbers=none]
{{#system~}}
You are a student and your teacher gives you an assignment to write a poem.
{{~/system}}

{{#user~}}
The assignment is:
{{assignment}}

{{#if exists_intrusction}}
In addition, here are some helpful advice and guidance:
{{instruction}}
{{/if}}
{{~/user}}
\end{verbatim}
\end{small}

For the \textbf{feedback synthesizer module}, we use this prompt:
\begin{small}
\begin{Verbatim}[breaklines=true]
{{#system~}}
You are a helpful assistant who aims to provide feedback to a student who's writing a poem 
according to some instructions.
It is important to let the student know if they did satisfy the instruction or not and why.
{{~/system}}

{{#user~}}
This is the history of past generated poems and how well they did with respect to instructions.

{{#each history}}
Instruction: {{this.observation}}

Poem: 
{{this.action}}

Feedback from the teacher: 
{{this.feedback}}
---------------
{{~/each}}
{{~/user}}

{{#user~}}
Now, the student writes a new poem.

New instruction: {{observation}}

Poem: 
{{action}}

What changes can you make to the poem to help it conform to the instructions?
{{~/user}}

{{#assistant~}}
{{gen 'exp_feedback' temperature=0.7}}
{{~/assistant}}
\end{lstlisting}
\end{small}

For the \textbf{LLM-based optimizer}, we use this prompt:

\begin{small}
\begin{lstlisting}[language=HTML,breaklines=true,showstringspaces=false,basicstyle=\fontsize{7}{7}\ttfamily, numbers=none]
{{#system~}}
You are a helpful assistant that wants to come up with instructions to a student to help them write a poem that is satisfactory to a teacher's assignment.
The student's poem needs to satisfy the requirement of this assignment.
{{~/system}}

{{#user~}}
This is the history of how you have been helping this student and whether your instructions have succeeded.
Teacher's feedback is the most important feedback, because the student needs to meet the teacher's criteria.
However, another student's feedback can provide helpful information too.

{{#each history}}
The Assignment: "{{this.assignment}}"

Your Instruction: 
{{this.prompt}}

Student's Poem: 
{{this.action}}

Teacher's Feedback: 
{{this.feedback}}

Feedback from another student:
{{this.exp_feedback}}
---------------
{{~/each}}
{{~/user}}

{{#user~}}

Your previous instruction didn't work -- the students didn't write a poem that satisfied the teacher's criteria.
Based on your interaction with the students, can you come up with better instructions that can help this student write a poem that matches the teacher's criteria?
Keep in mind that telling the student what to do step-by-step might be very helpful!
However, you need to be brief and to the point.
{{~/user}}
\end{lstlisting}
\end{small}

\subsection{Implementation Detail of the Tasks}

For both tasks, the implementation details can be found in \cite{cheng2023llf}, and the code is available at \url{https://github.com/microsoft/LLF-Bench}.

\end{document}